\documentclass[conference,10pt]{IEEEtran}

\renewcommand{\thesection}{\arabic{section}}

\usepackage[utf8]{inputenc}
\usepackage[T1]{fontenc}
\usepackage{textcomp}
\usepackage{fixltx2e}
\usepackage{multirow}
\usepackage{algorithm}
\usepackage{algpseudocode}
\usepackage[pdftex]{graphicx}
\DeclareGraphicsExtensions{.pdf,.png,.jpg}
\usepackage[caption=false,font=footnotesize]{subfig}
\usepackage{booktabs}
\usepackage{url}
\usepackage{amsmath}
\usepackage{amssymb}
\usepackage{amsthm}
\usepackage[capitalise,noabbrev]{cleveref}
\usepackage[textwidth=1cm]{todonotes}
\usepackage{xspace}
\usepackage{booktabs}
\usepackage[textwidth=1cm]{todonotes}
\usepackage[american]{babel}
\usepackage[color]{changebar}
\usepackage{textgreek}
\usepackage{enumitem}
\usepackage[bottom]{footmisc}
\usepackage{stfloats}
\usepackage{mathrsfs}

\usepackage[load-configurations=binary,detect-all,binary-units=true,range-phrase=--,per-mode=symbol,range-units=single,list-units=single]{siunitx}
\DeclareSIUnit\dBm{dBm}
\DeclareSIUnit\dB{dB}

\RequirePackage{xstring}
\RequirePackage{xparse}
\RequirePackage[]{acro}
\NewDocumentCommand\acrodef{mO{#1}mG{}}{\DeclareAcronym{#1}{short={#2}, long={#3}, #4}}
\acrodef{IoT}{internet of things}
\acrodef{V2X}{vehicle-to-everything}
\acrodef{ITS}{intelligent transportation system}{short-plural=}
\acrodef{ATIS}{advanced traveler information system}{short-plural=}
\acrodef{AVCS}{advanced vehicle control system}{short-plural=}
\acrodef{ATMS}{advanced traffic management system}{short-plural=}

\newcommand{\sv}{\mathbf{s}}

\IEEEoverridecommandlockouts

\title{\LARGE \bf Adaptive Autopilot: Constrained DRL for Diverse Driving Behaviors}

\author{Dinesh Cyril Selvaraj, Christian Vitale, Tania Panayiotou, Panayiotis Kolios, \\Carla Fabiana Chiasserini, and Georgios Ellinas
\thanks{Dinesh Cyril Selvaraj and Carla Fabiana Chiasserini are with CARS@Polito and Politecnico di Torino, Torino, Italy (e-mail:{\tt\small \{dinesh.selvaraj, carla.chiasserini\}@polito.it}). Christian Vitale, Tania Panayiotou, and Georgios Ellinas are with the KIOS Research and Innovation Center of Excellence and the Department of Electrical and Computer Engineering, University of Cyprus, Nicosia, Cyprus. Panayiotis Kolios is with the KIOS Research and Innovation Center of Excellence and the Department of Computer Science, University of Cyprus, Nicosia, Cyprus. (e-mail:{\tt\small \{vitale.christian, panayiotou.tania, pkolios, gellinas\}@ucy.ac.cy})}
\thanks{This work was supported by the EU’s Horizon 2020 research and innovation programme under grant agreement No 739551 (KIOS CoE - TEAMING) and under grant agreement No. 101069688 (CONNECT project), and from the Republic of Cyprus through the Deputy Ministry of Research, Innovation and Digital Policy.}}

\begin{document}

\maketitle
\thispagestyle{empty}
\pagestyle{empty}

\begin{abstract}
In pursuit of autonomous vehicles, achieving human-like driving behavior is vital. This study introduces adaptive autopilot (AA), a unique framework utilizing constrained-deep reinforcement learning (C-DRL). AA aims to safely emulate human driving to reduce the necessity for driver intervention. Focusing on the car-following scenario, the process involves (i) extracting data from the highD natural driving study and categorizing it into three driving styles using a rule-based classifier; (ii) employing deep neural network (DNN) regressors to predict human-like acceleration across styles; and (iii) using C-DRL, specifically the soft actor-critic Lagrangian technique, to learn human-like safe driving policies. Results indicate effectiveness in each step, with the rule-based classifier distinguishing driving styles, the regressor model accurately predicting acceleration, outperforming traditional car-following models, and C-DRL agents learning optimal policies for human-like driving across styles.
\end{abstract}

\section{Introduction}
In recent years, the automotive industry has experienced a digital transformation, enhancing vehicles with sensing devices, electronic control units, and advanced driver assistance algorithms, including features like blind-spot detection and adaptive cruise control (ACC) \cite{yurtsever_survey_2020}. This evolution aims to improve safety, traffic efficiency, and the overall travel experience \cite{yu_researches_2022}. However, consumer adoption relies on trust in automated systems and considerations on legal issues \cite{kyriakidis_public_2015}. 

The acceptance of these systems is also influenced by their ability to emulate human-like driving styles \cite{ma_drivers_2021, ma_investigating_2020}. Toward this, distinct driver categories, like aggressive drivers prioritizing smaller gaps with abrupt maneuvers and conservative drivers favoring larger gaps with smoother behavior \cite{sagberg_review_2015}, require tailored controllers. Current car-following controllers \cite{kesting_agents_2008, Treiber2013, etde_627062, gipps_behavioural_1981}, despite attempts to differentiate among driving styles, depend on predefined parameters, lacking real-world adaptability. The remedy lies in data-driven controllers utilizing real-world data, emulating diverse driving styles, and, theoretically, having the potential to reduce disengagement rates \cite{ma_drivers_2021}. In this context, machine learning (ML) plays a pivotal role in developing models capable of making informed decisions by analyzing complex and multi-variate data. Among ML paradigms, reinforcement learning (RL) is well-suited for this intricate task \cite{kiran_deep_2022, selvaraj_ml-aided_2023}, as RL agents learn by interacting with the environment through a trial-and-error mechanism, aiming to maximize cumulative rewards. However, traditional RL agents often overlook safety constraints critical for real-world applications like autonomous driving. C-DRL addresses this limitation as, unlike traditional RL, it incorporates constraints through cost functions, ensuring safe driving by minimizing them during the learning process \cite{altman_constrained_2021}.

Building on C-DRL, our work introduces the adaptive autopilot (AA) framework. This framework employs a C-DRL approach to effectively accommodate diverse driving styles by integrating rewards based on a human-like acceleration predictor, alongside constraints to enforce a minimum headway among vehicles. The three main steps of the framework are: (i) categorizing real-world driving data from the highD dataset \cite{krajewski_highd_2018} into aggressive, normal, and conservative styles, (ii) training deep neural network-based regressors to predict human-like vehicle acceleration tailored to each driving behavior, and (iii) implementing the C-DRL framework to take human-like safe actions. The trained agents, corresponding to each driving style and based on the soft actor-critic Lagrangian algorithm \cite{roy_direct_2021}, are validated using real-world driving data from the highD dataset. Results demonstrate the framework's ability to drive the vehicle in line with corresponding human drivers under different styles, with the headway trend highlighting the prioritization of safety constraints. To summarize, the main contributions of this work are as follows:

\begin{itemize}
\item[{\em (i)}] Real-world driving data is classified into aggressive, normal, and conservative styles using a rule-based approach. Separate neural network-based regressors are then trained for each style to predict human-like vehicle accelerations.

\item[{\em (ii)}] A novel C-DRL framework is introduced, adapting vehicle acceleration to different driving styles by (safely) mimicking human drivers. This is achieved through minimizing the difference between C-DRL and predicted human actions at each step. Further, a headway-based safety constraint is imposed during training, where multiple real-world driving traces are used to enhance generalization.

\item[{\em (iii)}] Performance results demonstrate the proposed framework's ability to adapt to diverse driving styles while adhering to safety constraints.
\end{itemize}

In the remainder of the paper, Sec. \ref{sec:relwrk} describes related research, Sec. \ref{sec:methodology} introduces the AA framework, Sec. \ref{sec:peval} discusses the obtained results, and Sec. \ref{sec:conc} presents concluding remarks.

\section{Related work \label{sec:relwrk}}
While commercial ACC systems were introduced in the early 2000s to enhance safety and driving experience \cite{watanabe_development_1995}, they still offer limited customization options with few user-defined parameters like desired gap and velocity. The rigidity of these systems hampers their ability to accurately replicate human driving styles, leading to reduced trust and increased driver intervention, thereby affecting safety benefits. Various research directions \cite{zhang_car-following_2023, papathanasopoulou_towards_2015, selvaraj_ml-aided_2023} have been explored to address these limitations and enhance ACC systems.

One research direction involves car-following (CF) models to provide optimal control actions in response to lead vehicle movements. Relevant models include the Gipps model \cite{gipps_behavioural_1981}, which prioritizes a safe inter-vehicle distance, incorporating human factors like reaction time and comfort. The intelligent driver model (IDM) \cite{Treiber2013} considers desired velocity and inter-vehicle distance, using different parameter values for various driving styles \cite{kesting_agents_2008}. However, these CF models struggle to accurately represent real-world driving behavior due to oversimplification, and their parameters are calibrated for traffic scenarios and safety rather than human-like driving behavior \cite{zhang_car-following_2023}. Our framework, compared to IDM, employs a data-driven approach demonstrating safe and human-like acceleration behavior across different styles.

Another research direction explores data-driven models, optimizing vehicle control using real-world mobility traces. For example, \cite{papathanasopoulou_towards_2015} employs  particle swarm optimization with bi-directional long short-term memory (PSO–Bi–LSTM) to enhance IDM model parameters and predict human driving behavior. IDM's learned fixed parameters limit however its ability to accurately model driving behavior. Other works use traditional and recurrent neural networks (NNs) for acceleration/velocity predictions \cite{khodayari_modified_2012, wang_capturing_2018}. Such NNs face challenges in personalized driver behavior modeling due to training data influences. Similarly, DRL has been utilized \cite{zhu_safe_2020, selvaraj_ml-aided_2023} for improved car-following behavior, emphasizing safety, traffic efficiency, and comfort. Nevertheless, these DRL approaches focus solely on generic driving behaviors, lacking consideration for a human in their training process to achieve human-like driving behavior.

Finally, the offline human-in-the-loop RL paradigm gains popularity for enhancing RL frameworks' adaptability by incorporating the human factor \cite{liang_human---loop_2017}. This approach does not require real-time human intervention but leverages human experience to shape reward functions. For instance, \cite{zhu_human-like_2018} uses the Shanghai naturalistic driving study data to mimic human-like driving behavior by designing reward functions to reduce errors between simulated and empirical values in spacing and velocity. It outperforms traditional NN models in capturing driver behavior, although safety concerns arise as aggressive human behaviors are replicated without considering safety. Additionally, \cite{tian_learning_2022} employs behavior cloning, an imitation learning method, to achieve human-like driving behavior. A major drawback of imitation learning is the accumulation of errors over time, leading to adverse control actions.

To the best of our knowledge, our work is the first to present a comprehensive human-in-the-loop C-DRL framework designed to adapt vehicle driving behavior across diverse driving styles along with safety constraints. 

\section{Adaptive Autopilot Framework\label{sec:methodology}}
Our framework addresses three interconnected problems: (i) identifying and classifying the driver's style using a rule-based approach based on headway, lead vehicle relative velocity, and acceleration (Sec. \ref{subsec:classify}); (ii) training a neural network-based regressor to predict human-like control actions, particularly acceleration rates, of the same driving style (Sec. \ref{subsec:reg}); (iii) implementing a C-DRL framework for the controller, considering vehicle states as input and ensuring safety while minimizing the difference between the control action and human-like acceleration predicted by the regressor (Sec. \ref{subsec:cdrl}).

\subsection{Rule-based Classifier}
\label{subsec:classify}
Inspired by \cite{kesting_agents_2008}, we categorize driving styles into aggressive, normal, and conservative. Such classification typically utilizes indicators related to longitudinal movements, including speed, acceleration, headway, relative velocity, as well as steering input and lateral acceleration  \cite{sagberg_review_2015}. Nevertheless, given the focus on car-following scenarios, only longitudinal control-related indicators are employed in this work. 

Designing a model to accurately classify a driver's entire data trace into a unique driving style is challenging due to potential variations within a driver's behavior. For instance, aggressive drivers may exhibit normal or conservative driving at times. In this work, each control action of the driver is tagged with a specific driving behavior. Subsequently, the ratio of each tagged behavior across the entire trace is calculated to categorize drivers as aggressive, normal, or conservative. Driver actions are tagged with a specific driving behavior based on a rule-based approach, which utilizes headway trends as a key factor in differentiating driving styles. As suggested by \cite{kesting_agents_2008}, aggressive drivers aim to maintain a headway of 1 second or below, Normal drivers aim for headways of around 1.5 seconds, while conservative drivers aim for a headway of 1.8 seconds and above. Based on longitudinal indicators, the classifier's objective is to tag the driver's intention, analyzing how the driver's action will change the headway and toward which of the three goal headways it will lead in the long term.


\begin{figure*}
\centering
\includegraphics[scale=0.62]{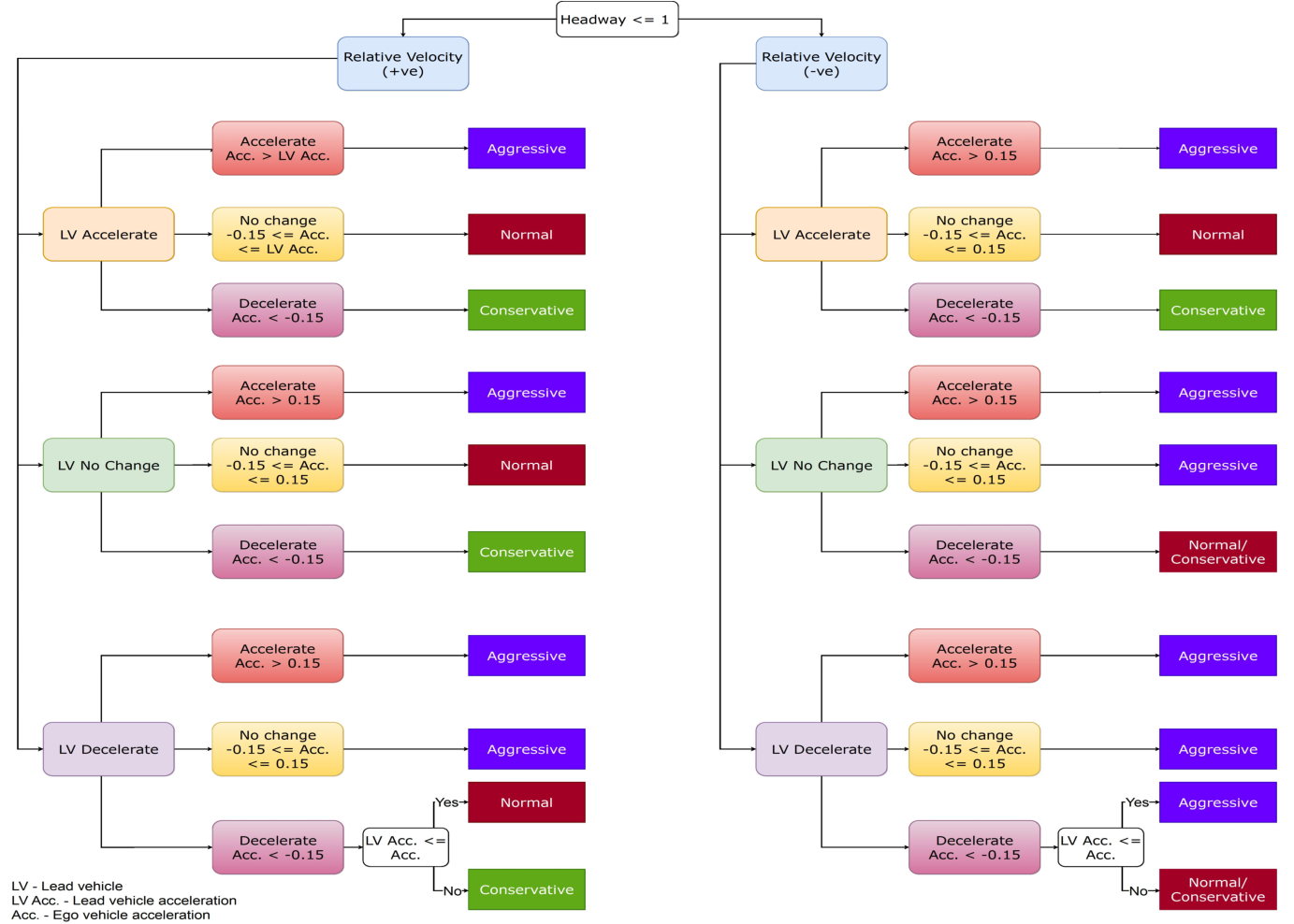} 
\vspace{-0.07in}
\caption{A hierarchical rule-based classifier (partial representation for the case of headway less than or equal to 1 sec) that labels the driving data into three driving style categories: Aggressive, Normal, and Conservative.}
\label{fig:rbClassifier}
\end{figure*}

Considering these aspects, Fig. \ref{fig:rbClassifier} outlines the hierarchical rule-based classification approach employed in this work, where the leaf nodes represent the assigned driving style. Note that only a partial classification is presented in Fig. \ref{fig:rbClassifier} (the case for headway less than or equal to 1 sec). Similar classifications have also been created for headway between 1-1.5 sec and for headway greater than 1.5 sec but are not presented here due to lack of space. Specifically, the classifier considers information related to both lead and ego vehicles at a given time $t$: $X(t) \mathord{=} \{\vartheta(t), \nu(t), \ddot{x}_{ego}(t), \ddot{x}_{lead}(t)\}$, representing headway, relative velocity, ego vehicle acceleration, and lead vehicle acceleration, respectively, to classify the behavior: $y(t) \mathord{=} \{Aggressive, Normal, Conservative\}$. Headway and relative velocity are formulated as:

\vspace{-0.08in}

\begin{eqnarray}
\label{stateEqn}
&& \vartheta(t) \mathord{=}  \frac{\Delta x_{lead}(t)}{\dot{x}_{ego}{(t)}}\, \label{hwCalc} \\ 
&&  \nu(t) \mathord{=} \dot{x}_{lead}{(t)} \mathord{-}  \dot{x}_{ego}{(t)},\label{rvCalc}
\end{eqnarray}  

\vspace{-0.08in}

\noindent where $\Delta x_{lead}(t)\mathord{=}x_{lead}(t)\mathord{-} x_{ego}(t)$ is the relative distance between lead and ego vehicles, and $\dot{x}_{lead}(t)$ and $\dot{x}_{ego}(t)$ represent the lead and ego vehicle's speed, respectively.

Among the input features, headway serves as the primary criterion for splitting the data into three leaf nodes. Subsequently, for each leaf node, relative velocity becomes a crucial factor in anticipating potential headway changes in subsequent time steps, acting as a secondary criterion for data segmentation. Considering the current headways and relative velocities, the driver's action, specifically the applied acceleration, is categorized into one of the three driving styles. Acceleration differences among the lead and ego vehicles enable an understanding of future changes in relative velocity before they manifest in the data points. This ability facilitates the identification of future achieved headways. Matching future achieved headways with desired headways (as per \cite{kesting_agents_2008}) provides a straightforward mean to categorize driving actions.

\subsection{Human-like Action Predictor}
\label{subsec:reg}
The human-like action predictor operates as a regressor model, using relevant input data to forecast the vehicle's acceleration. Its purpose is to learn a non-linear function that approximates the relationship between input data and the next vehicle's acceleration. To circumvent relying on past human actions, which might be unavailable in autopilot scenarios as the one of the AA framework, the regressor incorporates both historical and current data related to the lead vehicle, while utilizing only the present ego vehicle data as input. The input dataset consists of the following set of observations:
\vspace{-0.05in}
\begin{equation}
\begin{split}
X(t)\mathord{=}\{\ddot{x}_{lead}{(t\mathord{-} 2\Delta t)}, \ddot{x}_{lead}{(t\mathord{-}\Delta t)}, \ddot{x}_{lead}{(t)},\\ \dot{x}_{lead}{(t\mathord{-}2\Delta t)},\dot{x}_{lead}{(t\mathord{-}\Delta t)},\dot{x}_{lead}{(t)}, \dot{x}_{ego}{(t)}, \vartheta(t)\},
\end{split}
\end{equation}

\noindent to obtain prediction $\hat{y}(t)\mathord{=}\ddot{x}_{preg}{(t)}$ corresponding to the ego vehicle acceleration $y(t)\mathord{=}\ddot{x}_{ego}{(t)}$, where $t$, $t-\Delta t$, and $t-2\Delta t$ represent the present and two past time instants, respectively.

In this work, a DNN-based regressor, a deep learning technique, is utilized to predict human-like acceleration values. The highD dataset, \cite{krajewski_highd_2018}, serves as the training dataset, following the segmentation into the three driving styles mentioned above, achieved through the rule-based classifier outlined in Sec. \ref{subsec:classify}. Hence, a separate model is obtained for each driving style. Throughout the training process, the model is optimized to minimize the mean absolute error (MAE) loss function:
\vspace{-0.05in}
\begin{equation}
L_{mae}=\frac{1}{N} \sum_{i=1}^N|{y}_{i} -  \hat{y}_{i}|,
\label{mae_loss} 
\end{equation} 

\noindent where $N$ is the number of observations used for MAE loss minimization, $\hat{y}_{i}$ is the predicted value of the $i^{th}$ observation, and $y_{i}$ is the actual value of the $i^{th}$ observation.

\subsection{C-DRL Framework}
\label{subsec:cdrl}
\begin{figure}
\begin{center}
\includegraphics[width=\linewidth]{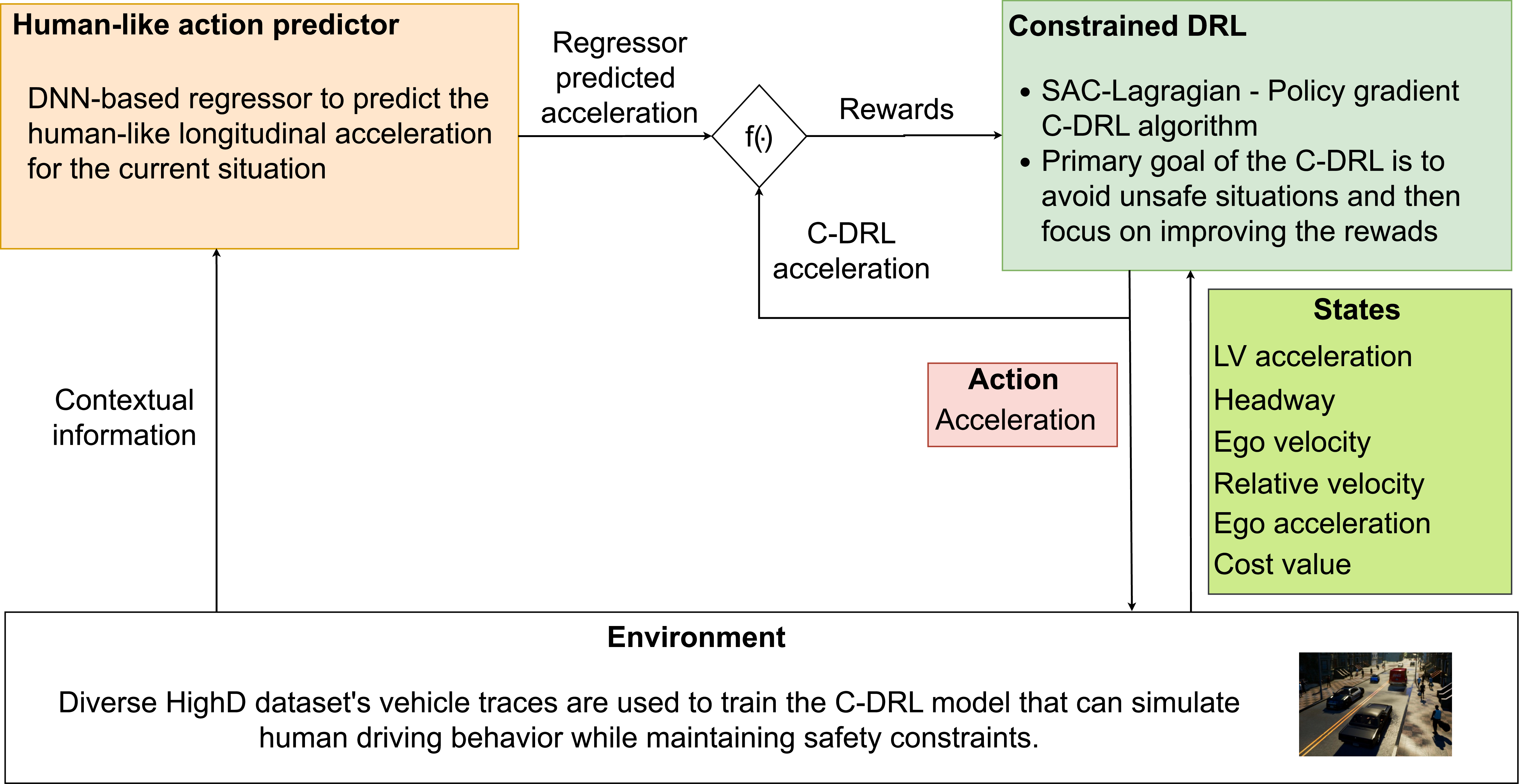}
\caption{An overview of the proposed AA methodology.} 
\label{fig:adapauto}
\vspace{-6mm}
\end{center}
\end{figure}

We now introduce our C-DRL framework, inspired by a previously proposed algorithm \cite{roy_direct_2021}. The C-DRL framework utilizes pertinent vehicle data as input to decide the vehicle's longitudinal control action, focusing specifically on vehicle acceleration. The control action applied guides the vehicle, earning rewards based on its ability to emulate the desired human-like driving behavior. Moreover, the framework integrates safety constraints to guarantee that the applied control actions maintain a safe distance between vehicles. Figure \ref{fig:adapauto} provides an overview of the proposed AA framework.

\noindent {\em \textbf{Background:}}
Constrained RL extends traditional RL by introducing constraints on the actions taken by the agent. C-RL is formalized as a constrained Markov decision process (CMDP) \cite{altman_constrained_2021}, an extension of the standard MDP framework. In this form, C-RL is characterized by the tuple $(\mathcal{S}, \mathcal{A}, \mathcal{P}, \mathcal{R}, \mathcal{C}, b, \gamma)$, representing the state space, action space, transition probabilities, reward, cost function, safety threshold, and discount factor, respectively. The goal of C-RL is to solve the CMDP by learning an optimal policy $\pi:\mathcal{S} \mathord{\rightarrow} \mathcal{A}$ that maximizes the expected cumulative discounted reward while satisfying the constraints. The problem addressed by C-RL is:

\begin{eqnarray}
\vspace{-0.06in}
\begin{aligned}
\label{eq:c_DRL}
&\max_{\pi:(\sv(t), a(t)) \sim \rho_{\pi}}\mathbb{E}\Bigg[\sum_{t} \gamma^t\mathcal{R}(\sv(t), a(t))\Bigg] \\   &\quad\quad\textrm{subject to:}\quad \mathbb{E}\Bigg[\sum_{t} \gamma^t\mathcal{C}(\sv(t), a(t))\Bigg] \leq b
\end{aligned}
\vspace{-0.05in}
\end{eqnarray}

\noindent where $\rho_{\pi}$ denotes the trajectory distribution following policy $\pi$, and $\mathcal{R}(\sv(t), a(t))$ and $\mathcal{C}(\sv(t), a(t))$ represent (resp.) the reward and cost functions associated with state $\sv(t)$ and action $a(t)$ at a specific time step $t$. As modeling the state transition probabilities for intricate problems can be challenging, in C-RL a model-free approach is typically used, with the relationship between action and reward/cost implicitly learned by interacting with the environment.

In constrained optimization problems, such as Eq. (\ref{eq:c_DRL}), C-RL can utilize an equivalent formulation with Lagrangian multipliers ($\lambda$) for optimization. The Lagrangian's saddle point is determined through iterative gradient ascent steps for the policy function $\pi$ and gradient descent on the Lagrangian multipliers $\lambda$ \cite{ha_learning_2020, yang_wcsac:_2021, roy_direct_2021}. Notably, the gradient step related to $\lambda$ emphasizes the loss function associated with the constraint. If the constraint is violated, the gradient update increases the multiplier's value, prioritizing the constraint over the reward function, and vice versa. C-DRL advances upon C-RL by incorporating deep neural network-based function approximators to model the policy function $\pi(\sv|\theta)$, where $\theta$ denotes the neural network parameters. This augmentation significantly improves the C-RL framework's capability to navigate intricate, high-dimensional real-world environments. In this study, we specifically adopt the soft actor-critic Lagrangian (SAC-Lagrangian) technique \cite{roy_direct_2021} as the C-DRL methodology to achieve the desired outcome.

\noindent {\em \textbf{States and Action Space:}} 
The C-DRL state, $\mathbf{s}(t) \mathord{\in} \mathcal{S}$, represents the vehicle state at any time $t$ and is given by:

\vspace{-3mm}
\begin{equation}
\sv(t){=}\{\ddot{x}_{lead}{(t)}, \vartheta(t), \dot{x}_{ego}{(t)}, \nu(t), a(t\mathord{-}\Delta t), \psi(t)\}
\end{equation}

\noindent where $a(t\mathord{-}\Delta t)$ denotes the control action taken by the C-DRL framework at time $t\mathord{-}\Delta t$ and $\psi(t)$ represents the value obtained from the indicator cost function associated with the safety constraint. The cost function is formulated as:

\vspace{-3mm}
\begin{equation}
\psi(t)\mathord{=}I(\vartheta(t)>\omega),
\label{costFunction} 
\vspace{-0.04in}
\end{equation} 

\noindent where $\omega$ represents the safety threshold for the distance between the vehicles. As mentioned earlier, the action space, $a(t) {\in} \mathcal{A}$, corresponds to the vehicle's acceleration (a continuous variable bounded within the range $[-4, 4]\,$ms$^{-2}$). Additionally, consecutive acceleration values are restricted to vary by no more than $\pm0.24$\,ms$^{-2}$ \cite{salles_extending_2022}.

\begin{figure}[h]
\begin{center}
\vspace{-3mm}
\includegraphics[width=1\columnwidth]{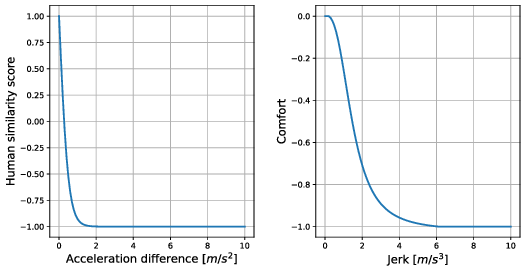}
\vspace{-4mm}
\caption{Human similarity (left) and comfort (right) reward trend.}
\label{fig:rewardtrend}
\vspace{-4mm}
\end{center}
\end{figure}

\noindent \textbf{\emph{Reward Components:}}
The reward signal is a scalar value provided by the environment after each action, offering insights into the agent's performance concerning the framework objectives. Our reward function comprises two key components: (i) human similarity reward, which assesses the disparity between C-DRL control actions and human-like actions predicted by the regressor model; and (ii) comfort, ensuring smooth acceleration changes between time steps. The trends of these reward components are illustrated in Fig. \ref{fig:rewardtrend}. Formally, the reward is expressed as:

\vspace{-3mm}
\begin{equation}
\label{eq:reward}
r(\sv(t),a(t)) \mathord{=}  r_{h}(\sv(t),a(t)) \mathord{+}  r_{c}(\sv(t),a(t)),
\vspace{-0.04in}
\end{equation}

\noindent where $r_{h}(\sv(t),a(t))$ and $r_{c}(\sv(t),a(t))$ represent human similarity and comfort rewards at time step $t$, respectively. The reward components are further detailed below.

\noindent \emph{Human Similarity Reward Component:} This reward component assesses the similarity between the driving behavior of the vehicle and that of a human. It quantifies the disparity between the predicted acceleration values by the DNN regressor and the ones applied by C-DRL, encouraging the agent to minimize this difference. Specifically, the function offers a reward that is maximum (+$1$) for zero error and decreases significantly as the difference between predictions increases. The reward formulation incorporates a \emph{$\tanh$} function for this purpose:
\setlength{\abovedisplayskip}{3pt}
\setlength{\belowdisplayskip}{3pt}
\begin{eqnarray}
&& r_{h}(\sv(t),a(t)) \mathord{=} 2 \cdot F_{h} \mathord{+} 1, \quad \mbox{with} \quad \\
&& F_{h} = \tanh(-2 \cdot \xi(t)), \quad \\
&& \xi(t) \mathord{=} |a(t) - \ddot{x}_{preg}{(t)}|.
\end{eqnarray}

\noindent \emph{Comfort Reward Component:} Sudden acceleration changes can lead to passenger discomfort. To address this, the comfort reward component considers the rate of change of acceleration with time, known as jerk ($j(t)$). The reward function is designed to decrease gradually as the absolute jerk value increases, ranging from a maximum reward value of $0$ to a minimum of $-1$. This desired reward trend is crafted using a curve-fitting function, specifically a 4PL model, as illustrated in Fig. \ref{fig:rewardtrend} (right), depicting the comfort reward trend.

\noindent \textbf{\emph{Simulation Environment:}}
A straightforward car-following simulation environment is created to replicate vehicle movements, enabling the C-DRL agent to learn the desired behavior. Utilizing the highD dataset, the simulation environment incorporates movements for the lead vehicle, while simulating the ego vehicle's motions using a linear motion model. The C-DRL agent's predicted acceleration serves as the control action to drive the vehicle, with a set sampling interval of $\Delta t$=$80$ ms. The ego vehicle movements follow:

\vspace{-3mm}
\begin{equation}
\begin{split}
\label{motionModel}
\dot{x}_{ego}{(t\mathord{+}\Delta t)} \mathord{=} \dot{x}_{ego}{(t)} \mathord{+} a(t) \Delta t, \\
{x}_{ego}{(t\mathord{+}\Delta t)} \mathord{=} {x}_{ego}{(t)} + \dot{x}_{ego}{(t)} \Delta t \mathord{+} 0.5 a(t) {\Delta t^2}.
\end{split}
\end{equation}  

\noindent \textbf{\emph{Learning Process:}}
In C-DRL, the exploration-exploitation process is crucial, involving a balance between trying new actions and exploiting high-reward actions. Initial training stages necessitate thorough exploration of the action space to discover those maximizing cumulative rewards. However, if action values are restricted, hindering exploration, the agent may miss identifying actions leading to higher rewards. To mitigate this, we adopt a curriculum learning approach \cite{hacohen_power_2019, markudova_recoco:_2023}, gradually increasing difficulty during training. Initial episodes allow unrestricted changes in subsequent actions, with limits introduced once the agent learns the desired behavior. Training utilizes multiple driver traces from the highD dataset for each driving style to ensure generalization, and it continues until satisfactory and stable rewards are achieved.

\section{Performance Evaluation\label{sec:peval}}
In this section, we introduce the dataset we used for training and evaluation of our proposed framework, and present the performance of our solution.

\subsection{Dataset}
\label{subsec:dataset}
This work utilizes the highD dataset, comprising vehicle trajectories recorded via a drone on German highways at six locations, each covering $420$ meters \cite{krajewski_highd_2018}. The dataset consists of $110,500$ vehicle trajectories across $60$ recordings, with an average recording length of $17$ minutes, encompassing free-driving, car-following, and lane-changing events. To focus on the car-following scenario, vehicle traces were filtered based on criteria including duration (minimum $10$ s of data), absence of lane changes, consistent lead vehicle, minimum speed ($6$ ms$^{-1}$), and vehicle type classification. The sampling frequency used in this work is $\Delta t$=$80$ ms. Among the $60$ recordings, $32$ were selected for pre-processing, resulting in approximately $2.6$ million rows of data, balancing accuracy in training regressor models with computational efficiency.

\subsection{Performance Results: Rule-based Classifier \label{subsec:prRBC}}
\begin{figure}
\begin{center}
\includegraphics[scale=0.8]{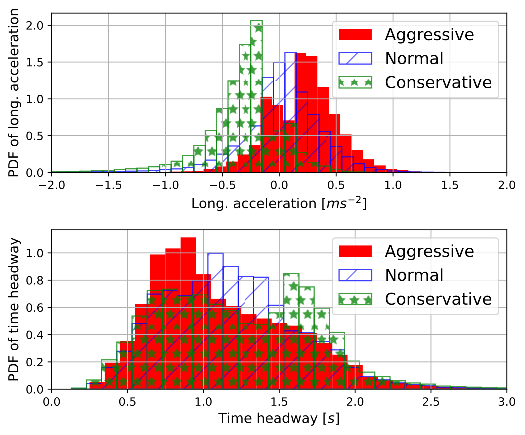}
\vspace{-4mm}
\caption{Analysis of driving styles classification using the highD dataset.}  
\label{catAnalysis}
\end{center}
\end{figure}

Here we showcase the performance of the rule-based classifier on the highD dataset. Based on the rules presented in Sec. \ref{subsec:classify}, the data points are classified into three categories: Aggressive, Normal, and Conservative. Figure \ref{catAnalysis} depicts the key characteristics, in terms of longitudinal acceleration and time headway, of each category using this rule-based setup. Overall, aggressive, normal, and conservative driving behaviors comprise 924k, 1.4M, and 863k data points, respectively, with some data double-tagged because certain behaviors coincide with more than one driving style.

The performance results of the classifier are consistent with expectations, showing large differences between driving styles. Looking at the probability density function (PDF) (Fig. \ref{catAnalysis} (top)) of the applied longitudinal acceleration, conservative drivers tend to brake to increase the distance from the lead vehicle. Specifically, the PDF mode is at $-0.2$ ms$^{-2}$ and $85\%$ of the conservative actions represent a braking action. On the contrary, aggressive drivers aim to close the gap with the lead vehicle as much as possible (PDF mode is at 0.2 ms$^{-2}$, and $73\%$ of the aggressive actions represent acceleration). Normal driving follows a hybrid pattern, with the PDF mode at around 0 ms$^{-2}$.

Further, although classification happens based on the driver's intention to change its headway, the headway PDF plots (Fig. \ref{catAnalysis} (bottom)) show that the mode behavior of a specific driving style corresponds to the ones envisioned in \cite{kesting_agents_2008} (i.e., the PDF mode of aggressive drivers just below the 1-s mark, of normal drivers between 1 and 1.5 s, and of conservative drivers just before the 2-s mark).

\subsection{Performance Results: Acceleration Prediction \label{subsec:PRDregModel}}
This section discusses the performance of the regressor models corresponding to the three driving behaviors. As mentioned earlier, we employed a traditional DNN regressor to train the models for predicting vehicle acceleration based on the input data. Using the categorized data obtained from the rule-based classifier, each driver behavior dataset was divided into training ($65\%$), validation ($15\%$), and testing ($20\%$) sets. Given the varied nature of the input, we employed a standardization technique to scale the input features, with zero mean and standard deviation equal to one to help the training. Also, to mitigate overfitting, an early stopping technique was employed to stop training if the error improvement was less than $0.001$ in the validation set for five epochs. It is noted that this section presents the best configurations for each model after extensive hyperparameter trials (Tab.\,\ref{tab:DNNhyperparameter}). 

\begin{table}[]
\centering
\caption{DNN regressor's hyperparameter values}
\vspace{-3mm}
\label{tab:DNNhyperparameter}
\begin{tabular}{cccc}
\hline
\multirow{2}{*}{}  & \multicolumn{3}{c}{Driving style} \\ \cline{2-4} 
& Aggressive & Normal & Conservative \\ \hline
Number of \\ hidden units (3 layers)   & 256, 128, 64  & 256, 256, 128 & 256, 128, 64    \\ \cline{1-4}  Dropout rate \\ between hidden layers  & 0.2,0.15,0.1 & 0.2,0.15,0.1 & 0.2,0.15,0.1 \\ \cline{1-4}  
Learning rate   & 0.0001 & 0.0001 & 0.0001     \\ \cline{1-4}  
Batch size   & 32 & 64 & 64    \\ \hline
\end{tabular}
\vspace{-4mm}
\end{table}

MAE was used to compare the results obtained during inference. In Tab. \ref{tab:maeError}, the results obtained by the proposed DNN regressor model are compared with the well-known car-following algorithm, IDM, with parameters suggested in \cite{kesting_agents_2008}. Additionally, the IDM model was enhanced to match as closely as possible the highD dataset. The fixed parameters used in \cite{kesting_agents_2008} were modified so as to obtain the best possible fitting with the dataset data points, i.e., the IDM fixed parameters were selected so that the MAE was minimized (IDM-MAE). The results obtained demonstrate that the proposed regressor models outperform the car-following algorithms for all driving styles. Further, analyzing the CDF of the MAE (not presented in the paper due to space limitations), the optimal performance of the DNN predictor is also showcased by the fact that the absolute error of the prediction, i.e., $|\ddot{x}_{ego}(t)\mathord{-}\ddot{x}_{pred}(t)|$, is less than $0.21$ ms$^{-2}$ in $80\%$ of the data points for all driving styles.

\begin{table}[h!]
\vspace{-3mm}
\centering
\caption{Mean absolute prediction errors}
\vspace{-3mm}
\label{tab:maeError}
\begin{tabular}{|c|c|c|c|}
\hline
Driving style & DNN & IDM & IDM-MAE\\ 
\hline
\textbf{Aggressive} & 0.1356 & 2.0357 & 0.3936\\ 
\hline 
\textbf{Normal} & 0.1413 & 2.4309 & 0.4584\\ 
\hline 
\textbf{Conservative} & 0.1415 & 4.3752 & 0.6151\\ 
\hline 
\end{tabular}
\vspace{-2mm}
\end{table}

\begin{figure}
\begin{center}
\includegraphics[width=\columnwidth]{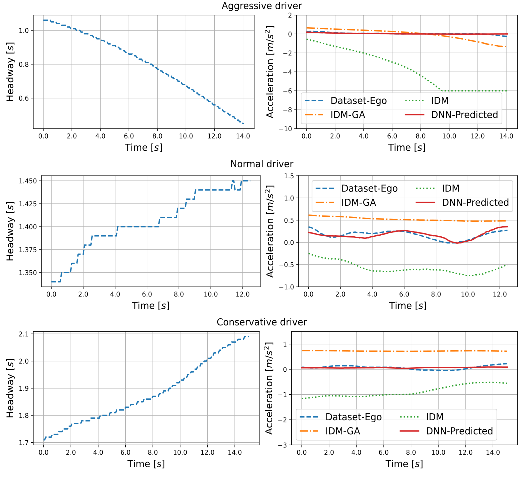}
\vspace{-8mm}
\caption{Driver-wise regressor model predictions for the three driving styles.}    
\vspace{-5mm}
\label{staticPlots}
\end{center}
\end{figure}

To assess the predictor's performance on individual driver traces, Fig. \ref{staticPlots} displays three traces representing aggressive (top), normal (middle), and conservative (bottom) driving behaviors. To evaluate their long-term performance, for both the DNN predictor and the benchmarks, the vehicles in the traces are moved by applying the predicted accelerations using the motion model in Eq. \ref{motionModel}. That is, while the lead vehicle traces correspond to those in the highD dataset, the ego vehicle trace disregards the driver's applied accelerations but incorporates the acceleration predicted by the DNN and benchmarks.

Despite some deviations, the DNN-based predicted acceleration closely matches the actual driver behavior and outperforms existing benchmarks. The slight discrepancies observed can be attributed to the DNN predictor leveraging data from thousands of drivers to learn how to predict acceleration in specific situations, while individual driver styles may vary slightly even within the same driving behavior. Although space constraints prevent us from presenting it, applying the wrong driving style's DNN regressor in Fig. \ref{staticPlots} would yield significantly different results, with vehicle headways consistently diverging from the true values experienced by the drivers.

\vspace{-3mm}
\begin{table}[h]
\centering
\caption{C-DRL hyperparameter values}
\vspace{-3mm}
\label{tab:CDRLhyperparameter}
\begin{tabular}{ccc}
\hline                
& SAC-Lagragian \\ \hline    
Number of hidden layers (actor, critic)   & 3, 2   \\ \hline
Number of hidden units (actor)   & 128, 256, 128   \\ \hline
Number of hidden units per layer (critic)   & 128   \\ \hline
Learning rate & 0.0003 \\ \hline
Replay buffer size & 1,000,000 \\ \hline    
Mini-batch size & 128 \\ \hline
Discount factor & 0.99 \\ \hline
Number of random exploration episodes & 100 \\ \hline
Number of transitions between updates & 5 \\ \hline
Constraint threshold & 0.1 \\ \hline
\end{tabular}
\end{table}
\vspace{-3mm}

\subsection{Performance Results: Human-like Driving \label{subsec:PR_CDRLl}}
This section discusses the results of the C-DRL models, aiming to mimic human behavior safely. One model was trained for each driving style, with hyperparameter values similar to those in SAC-Lagrangian \cite{roy_direct_2021} (except for the differences listed in Tab. \ref{tab:CDRLhyperparameter}). Diverse traces were used for each driving style to ensure learning generalized behavior. For each agent, a driving trace was randomly selected from the ones chosen for training for each episode.

\begin{figure}
\includegraphics[width=\columnwidth]{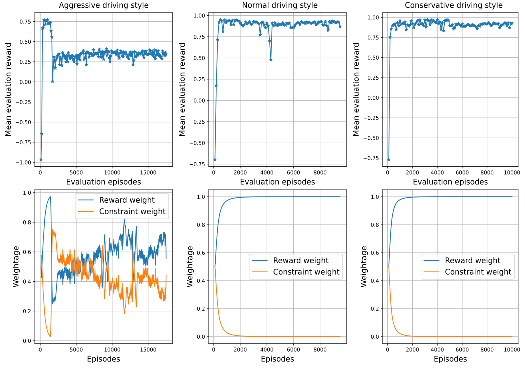}
\vspace{-7mm}
\caption{C-DRL training: reward trend (top) and reward vs constraint importance (bottom) for Aggressive, Normal, and Conservative driving.}
\label{learning}
\vspace{-3mm}
\end{figure}

The safety objective is to maintain a minimum headway of $\omega$=$1$ s between two vehicles. Hence, the final accelerations decided by the C-DRL agent must ensure that the two vehicles are never closer than this minimum headway. To achieve this objective, during training, the agent primarily focuses on finding a policy that minimizes the cost function representing the safety constraint. After ensuring the safety constraint is met, the agent tries to maximize the reward function, aiming to find a comfortable acceleration profile that mimics human-like driving behavior. Figure \ref{learning} presents the evolution during training of the rewards and the weights assigned to the cost ($\lambda$) and reward ($1$-$\lambda$) functions for the three agents. Specifically, the first row shows the reward trend of the evaluation episodes during training (executed every $100$ training episodes), which is used to assess the training progress. As depicted, the reward grows and stabilizes as training progresses. The normal and conservative agents achieve higher rewards than the aggressive agent because the aggressive agent prioritizes the safety constraint cost function before maximizing rewards (Fig. \ref{learning} (second row)). The conservative and normal driving behavior agents give more importance to the rewards, as the corresponding agents would not breach the safety constraint (according to their driving style). However, for aggressive drivers, who would naturally drive the headway below the 1-sec mark, the optimal cost function weight $\lambda$ is not equal to zero. This indicates that reward maximization would fail to respect the safety constraint, which is undesirable. Hence, the final agent trades reward maximization for enhanced safety. To test each agent's performance, we selected the best-performing model from the evaluation episodes for the inference phase.

\begin{figure}
\begin{center}
\includegraphics[width=1\columnwidth]{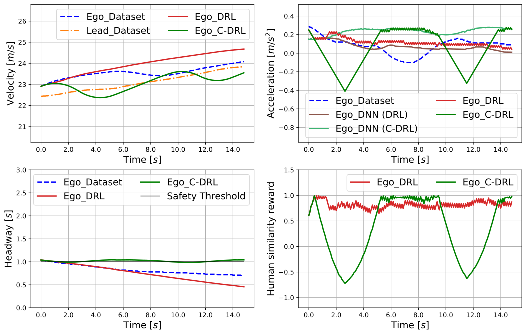}
\vspace{-8mm}
\caption{Inference phase: C-DRL - Aggressive driving style.} 
\vspace{-5mm}
\label{aggDriver}
\end{center}
\end{figure}

\begin{figure}
\begin{center}
\includegraphics[width=\columnwidth]{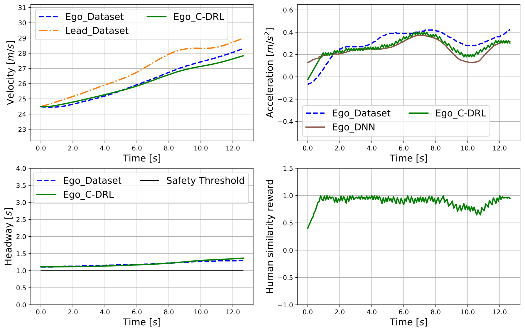}
\vspace{-8mm}
\caption{Inference phase: C-DRL - Normal driving style.}  
\vspace{-5mm}
\label{nrmlDriver}
\end{center}
\end{figure}

\begin{figure}
\begin{center}
\includegraphics[width=\columnwidth]{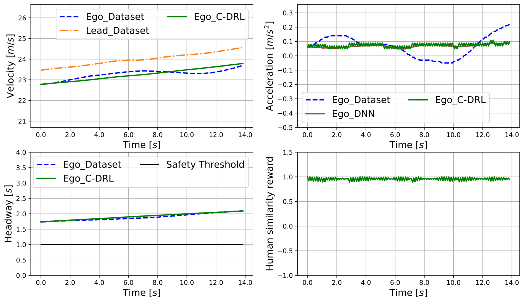}
\vspace{-8mm}
\caption{Inference phase: C-DRL - Conservative driving style.}   
\vspace{-5mm}
\label{consDriver}
\end{center}
\end{figure}

During inference, the agents are tested on driving traces that were not used during training. Figures \ref{aggDriver}$-$\ref{consDriver} illustrate the agent's performance across the three driving styles, each for a specific trace (with similar results obtained for all tested traces). Figure \ref{aggDriver} demonstrates that the aggressive agent could safely drive the vehicle, maintaining the headway around the 1-s mark without violating the safety threshold and also imitating the acceleration predicted by the regressor model whenever possible. When the headway drops below the safety threshold, the agent starts braking smoothly to maintain a safe distance from the lead vehicle. Once the agent has successfully satisfied the safety constraint, its focus shifts to mimicking the driver's behavior, as depicted by the acceleration trend (Fig. \ref{aggDriver} (top right)), closely following the DNN regressor model predictions. This is also confirmed by the human similarity reward trend (Fig. \ref{aggDriver} (bottom left)). To emphasize the importance of the C-DRL approach, we compared the proposed framework with a non-constrained DRL technique (referred to as Ego\_DRL), where we excluded the cost indicator function during training, confirming that without the safety constraint, the aggressive driving style could lead to unsafe headway, potentially resulting in dangerous situations.

In the normal and conservative driving styles, where the safety constraint's role is not crucial, the agents effectively mimicked human driving behavior by following the DNN regressor-predicted accelerations (Figs. \ref{nrmlDriver} and \ref{consDriver}). To quantitatively evaluate the results, we calculated the root mean square error between the regressor-predicted human-like acceleration and the C-DRL predicted acceleration, resulting in error values of $0.282$, $0.043$, and $0.013$ for aggressive, normal, and conservative driving behavior, respectively. As anticipated, aggressive driving behavior yields a higher error due to safety constraints, while errors for normal and conservative driving behaviors remain minimal. Additionally, it is noteworthy that, although slight discrepancies exist between the DNN regressor and the actual human-applied acceleration, the overall headway profiles generated by the C-DRL agents consistently align closely with those observed in the dataset.

\section{Conclusion\label{sec:conc}}
We presented an adaptive autopilot framework utilizing C-DRL to drive vehicles similarly to human drivers, adapting to diverse driving styles. The adaptive autopilot framework tackles three interconnected sub-problems: identifying driving styles using real-world data through a rule-based approach, predicting human-like acceleration across different driving styles using a DNN regressor model, and proposing a C-DRL approach to drive vehicles while considering safety constraints and mimicking human-like behavior. Results indicate that the regressor model can mimic human-like driving behavior as close and as safe as possible, and outperforms state-of-the-art IDM models in predicting acceleration. Hence, the comfortable experience provided by the proposed adaptive autopilot framework has the potential to enhance the satisfaction of human drivers, leading to a reduced disengagement rate of the autopilot driving system. Future work includes leveraging semi-supervised learning for enhanced driving style categorization and extending the framework to realistic environments that include complex scenarios like cut-ins and lane changes. 



\end{document}